\documentclass[10pt,twocolumn,letterpaper]{article}

\usepackage[pagenumbers]{cvpr} 

\usepackage{bm}
\usepackage{multirow}

\definecolor{cvprblue}{rgb}{0.21,0.49,0.74}
\usepackage[pagebackref,breaklinks,colorlinks,allcolors=cvprblue]{hyperref}

\def\confName{CVPR}
\def\confYear{2027}

\title{When to Start Denoising? \\Adaptive Denoising Entry for Generative Video Compression}
\author{
Zifu Zhang$^{1,2,3*}$,
Yuqiao Li$^{1*}$,
Ding Ding$^{2*}$,
Shengxi Li$^{1\dagger}$,
Mai Xu$^{1}$,
Daowen Li$^{2}$,
Kai Li$^{2}$,
Ying Chen$^{2\dagger}$ \\[6pt]
$^{1}$Beihang University \quad
$^{2}$Alibaba Group \quad
$^{3}$HKUST \\
}

\begin{document}
\maketitle
\begingroup
\renewcommand{\thefootnote}{\fnsymbol{footnote}}
\footnotetext[1]{Equal contribution.
\quad $^{\dagger}$Corresponding authors.}
\endgroup
\begin{abstract}
Generative video compression offers a promising solution for compressing and reconstructing videos under extremely low bitrate constraints, among which the diffusion-based solutions still achieve the best performance by providing the powerful generative latent prior. However, their effectiveness critically depends on when the compressed latent is introduced into the diffusion process, which directly governs the trade-off between reconstruction fidelity and perceptual realism. To address this limitation, we identify denoising start-timestep selection as a key factor governing this trade-off and formulate it as a geometric optimization problem. Based on this formulation, we propose a novel adaptive diffusion entry method for generative video compression (ADE-GVC), the first timestep-aware generative video compression framework, featuring a lightweight predictor that estimates an adaptive denoising start point from the compressed latent. A temporal adapter further extracts compression-aware conditions and injects them into a pretrained video diffusion model, enabling the reconstruction of videos with both high fidelity and realism. More importantly, we observe a strong quantitative relationship between the predicted timesteps and the distortion between the original and compressed latents, providing empirical support for our optimization formulation. Extensive experiments demonstrate that ADE-GVC achieves the state-of-the-art performance on perceptual metrics and substantially improves reconstruction realism over existing generative video compression methods. Notably, our ADE-GVC achieves an average bitrate savings of 40.99\% (on DISTS) and 95.24\% (on PSNR) compared to the latest open-source baseline GLC-Video. 
\end{abstract}

\section{Introduction}

\begin{figure}[ht]
    \centering
    \includegraphics[width=1.0\linewidth]{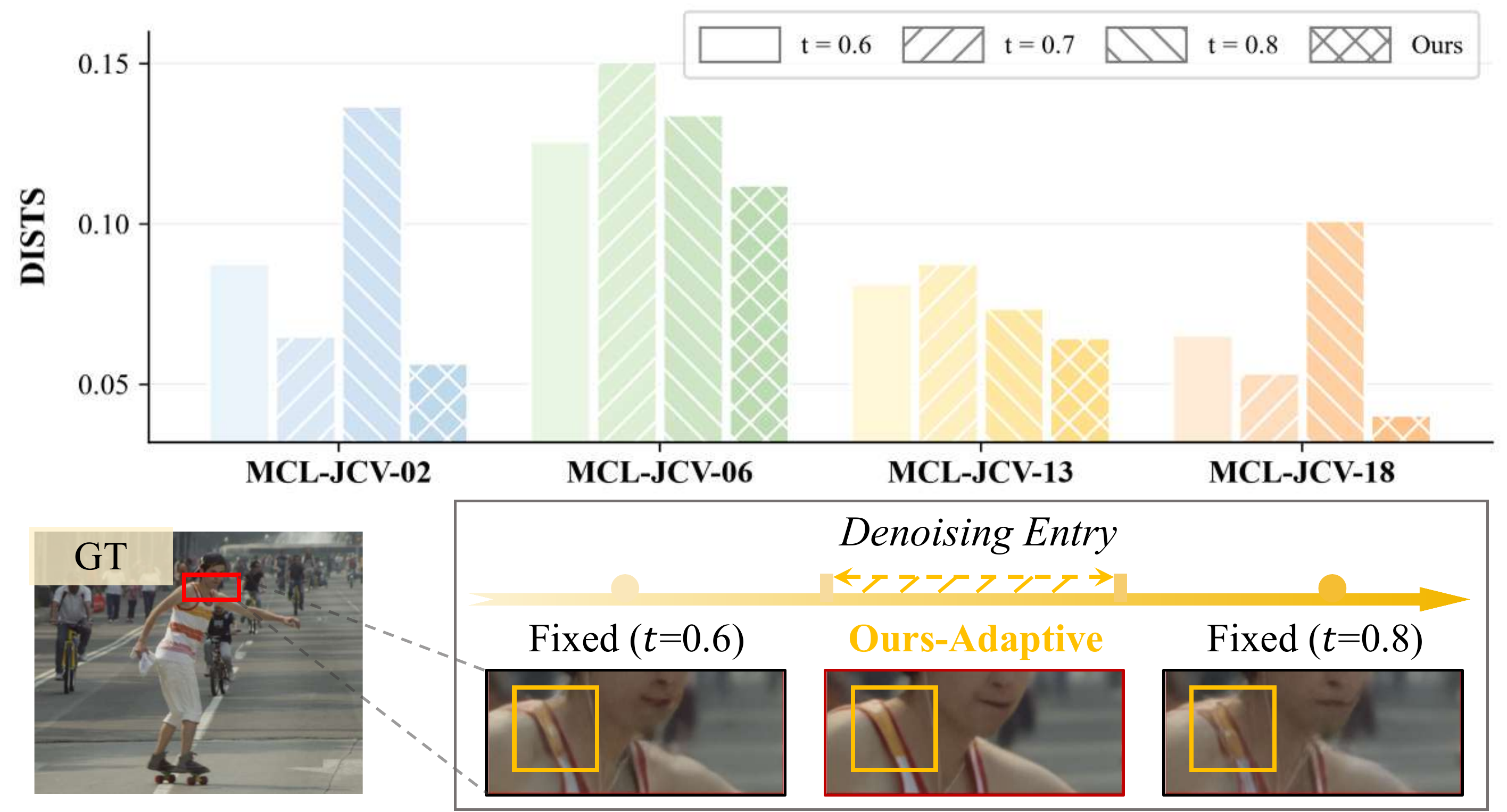}
    \caption{Illustration of the adaptive denoising entry scheme on MCL-JCV video sequences \cite{MCL-JCV}, in which fixed start timestep fails to restore various compression degradations. In contrast, we propose to optimize for adaptive entry timestep that is capable of addressing both fidelity and realism for generative video compression.}
    \vspace{-1em}
    \label{fig:intro}
\end{figure}

The rapid growth of video data has made efficient video compression increasingly important. Conventional video coding standards, including H.264/AVC \cite{wiegand2003overview}, H.265/HEVC \cite{sullivan2012overview}, and H.266/VVC \cite{bross2021overview}, have achieved remarkable coding efficiency. However, further improvements are becoming increasingly difficult, as these traditional hybrid frameworks rely on hand-crafted modules that are optimized separately. With recent advances in deep learning, learned video codecs such as DCVC series \cite{DCVC, DCVC-TCM, DCVC-HEM, DCVC-DC, DCVC-FM, DCVC-RT, DCVC-UF} have opened new avenues for improving compression performance. They jointly optimize the analysis and synthesis transform, entropy models and temporal context mining, introducing advanced spatio-temporal modeling. However, these methods are still predominantly optimized for objective distortion metrics such as mean squared error (MSE), which often produce over-smoothed and blurry frames that compromise visual quality for human vision. This problem is particularly severe at low bitrates, where overly quantized representations will be unable to support high-frequency details and textures.

In recent years, advances in image and video generation \cite{wan} have made the generative codec a promising paradigm for perceptual compression especially at low bitrates. {A generative codec, due to its ability to leverage the powerful prior knowledge in generative models, can allow for more aggressive coding strategies in rate-distortion optimization.} Video compression approaches based on generative adversarial networks (GANs) \cite{goodfellow2020generative} and diffusion models \cite{songdenoising} have demonstrated remarkable perceptual performance. GAN-based compression methods \cite{PLVC, GLC} offer relatively low inference complexity but remain limited in compression efficiency due to their insufficient generative capacity{, and their adversarial training strategy is often unstable which may introduce temporally inconsistent artifacts}. In contrast, diffusion-based methods \cite{DiffVC, S2VC, Mao_2026_CVPR, DiffVC-OSD, li2026generative} leverage powerful generative priors to substantially improve the perceptual realism of reconstructed videos. However, such improvements often realise at the cost of reconstruction fidelity, as the generated frames may deviate considerably from the original content and even introduce plausible but inaccurate details. Although prior studies \cite{pdtradeoff} have established a fundamental trade-off between distortion and perception, existing generative compression methods remain far from approaching the theoretical distortion-perception bound. Therefore, effectively exploiting generative priors while preserving faithful correspondence to the source video remains a critical challenge in low-bitrate video compression.

To address these limitations, we revisit the distortion–perception trade-off from the perspective of the diffusion denoising entry. Existing methods that leverage pretrained diffusion priors typically start denoising either from pure noise conditioned on the compressed representation \cite{li2024towards, ding2026controllable} or from the compressed latent perturbed with a fixed noise level \cite{li2025rdeic, Mao_2026_CVPR}. However, the injected noise inevitably disrupts source information and merely enables the usage of pretrained denoising priors. We argue that injecting a fixed noise level is suboptimal because it fails to accommodate variations in compression distortion across video sequences. Therefore, this paper formulates denoising entry selection as a geometric optimization problem and derives a closed-form solution under bounded model fine-tuning capacity. The analysis reveals that the optimal entry timestep decreases monotonically as compression distortion increases, indicating that more severely distorted latents require an earlier entry into the denoising process with a higher level of injected noise. Based on this formulation, we propose a novel \underline{A}daptive \underline{D}iffusion \underline{E}ntry method for \underline{G}enerative \underline{V}ideo \underline{C}ompression (ADE-GVC), the first timestep-aware generative video compression framework. It features a lightweight predictor that estimates an adaptive denoising start point from the compressed latent. Interestingly, the predicted timesteps exhibit a strong quantitative relationship with the distortion evaluated by DISTS \cite{DISTS}, empirically validating our theoretical solution. Videos with larger compression-induced distortion consistently select earlier denoising entries, even though the predictor receives no explicit supervision on distortion. Finally, a temporal adapter injects compression-aware conditions into the pretrained VACE \cite{vace} model, enabling faithful and realistic video reconstruction. As illustrated in Fig.~\ref{fig:intro}, different video sequences favor different discrete denoising entry points. In contrast, our method adaptively selects the optimal entry for each sequence, consistently achieving the best performance across all sequences. Extensive experiments demonstrate that ADE-GVC achieves state-of-the-art performance on perceptual metrics and substantially improves reconstruction realism over existing generative video compression methods. Our contributions are summarized as follows:
\begin{itemize}
\item We formulate the distortion–perception trade-off from the perspective of the denoising entry and validate its connection to the widely applied metric empirically.
\item We propose the first adaptive diffusion entry method for generative video compression, namely, ADE-GVC, overcoming the limitations of fixed-entry strategies.
\item ADE-GVC achieves the state-of-the-art performance, delivering average bitrate savings of 40.99\% in terms of DISTS and 95.24\% in terms of PSNR over the latest open-source baseline GLC-Video \cite{GLC}.
\end{itemize}

\section{Related Works}

\subsection{Learned Video Compression}


Neural video compression (NVC) algorithms~\cite{DVC, DCVC} have advanced rapidly and have in some cases surpassed conventional hybrid codecs in rate-distortion performance. Existing NVC methods are generally categorized based on their mechanism for reducing temporal redundancy, falling into either residual coding or conditional coding paradigms. Residual coding methods~\cite{DVC, HLVC, RLVC, PLVC, ALVC} explicitly encode inter-frame residuals. In contrast, conditional methods~\cite{DCVC, DCVC-TCM, DCVC-HEM, DCVC-DC, DCVC-FM, DCVC-RT} encode frames conditioned on the temporal context of the reconstructed references. These NVC methods are primarily devoted to optimizing reconstruction fidelity, which leads to excessive blurring of details under low-bitrate conditions.


Generative video compression~\cite{guo2025generativelatentvideocompression, GLC, GVC1D} addresses this limitation by leveraging generative priors and perceptual metrics to produce reconstructions with realistic textures and fine details. Early work~\cite{PLVC, GLC} employed GANs to improve perceptual quality, while more recent methods~\cite{DiffVC, DiffVC-OSD} have turned to diffusion and flow-matching models for their stronger generative capability. For instance, GNVC-VD~\cite{Mao_2026_CVPR} has introduced VideoDiT to enhance temporal consistency. Nevertheless, diffusion models require multiple iterative steps, which severely impairs decoding speed. S$^2$VC~\cite{S2VC} alleviates this problem by adopting a one-step diffusion model. These methods demonstrate the considerable potential of diffusion models for video compression. However, their use of the fixed timestep schedule for different video sequences may result in insufficient reconstruction fidelity. In contrast, our ADE-GVC adaptively determines the denoising depth of the diffusion model according to the properties of each sequence, thereby achieving better fidelity and realism.

\subsection{Adaptive Denoising Entry for Low-level Vision}


Although fixed denoising entries are common in generative compression, adaptive strategies have been explored in other low-level vision tasks, including super-resolution, enhancement, dehazing, and artifact removal~\cite{RPD-Diff, Hero-SR, duan2026fixedinferencequantitativeflow}. These methods typically estimate degradation severity or map inputs to explicit noise levels. For example, DMID~\cite{DMID} directly estimates the noise strength during noise-distribution transformation, while SODiff~\cite{yang2025sodiff} selects the corresponding timestep according to the JPEG quality factor. TADM~\cite{TADM_Wang_2025_ICCV} employs a neural network to estimate the noise level for image rescaling. Similarly, UltraVSR~\cite{ultraVSR} combines network-based noise-level estimation with noise scheduling to achieve one-step reconstruction from low-quality to high-quality videos.


However, compression differs fundamentally from other low-level vision tasks. \textit{Its degradation is not caused by the external environment but by a compression module, which is an internal distortion source and is jointly optimized within the network.} Consequently, compression demands substantially higher reconstruction fidelity, rendering existing adaptive methods inapplicable. Existing generative compression methods~\cite{Mao_2026_CVPR, li2025rdeic} typically adopt a fixed denoising entry by injecting a predefined noise level into the compressed representation. Although this design facilitates joint optimization with the compression network, the fixed noise injection may unnecessarily corrupt the retained source information and thereby limit the reconstruction performance of the diffusion model.

\section{Methodology}

\subsection{Problem Definition}

\begin{figure}[t]
    \centering
    \includegraphics[width=1.0\linewidth]{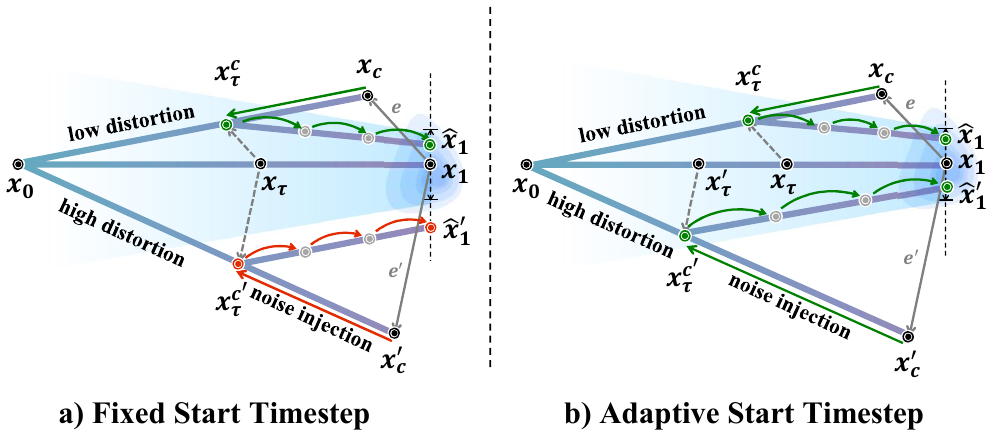}
    \caption{Theoretical analysis on our adaptive start timestep scheme against the fixed start timestep scheme, from the geometry perspective. The lightblue regions denote the bounded adaptation capacity of the pretrained model. We illustrate two cases, namely, denoising from the low-distortion latent $\bm{x}_c$ and high-distortion latent $\bm{x}'_c$. From the theoretical perspective, we can conclude that a fixed denoising entry cannot accommodate varying distortion levels, whereas our adaptive strategy can.
   }
    \label{fig:qd}
\end{figure}

Existing generative approaches to low-level vision tasks commonly build upon pretrained diffusion models. These approaches typically fine-tune the generative prior by incorporating degraded observations as conditioning signals and optimizing the resulting framework in an end-to-end manner, thereby improving reconstruction quality. However, these methods overlook the varying generative capabilities at different timesteps and are trained with a fixed timestep. To demystify this, we first formulate timestep selection as a geometric optimization problem, as illustrated in Fig.~\ref{fig:qd}.

\begin{figure*}[ht]
    \centering
    \includegraphics[width=1.0\linewidth]{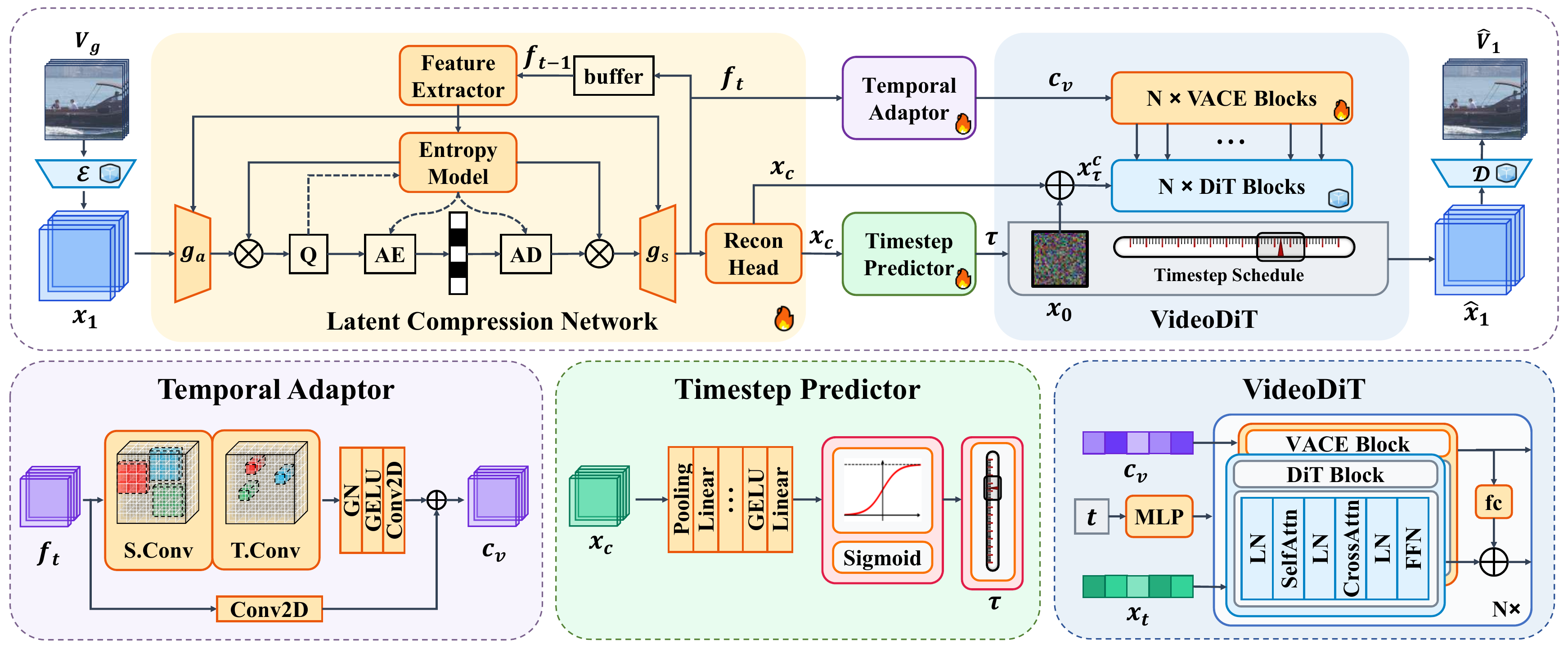}
    \caption{The overall architecture of ADE-GVC consists of a latent compression network (LCN), temporal adaptor (TA), timestep predictor (TP), and video diffusion model. The input video $\bm{V}_g$ is encoded into the clean latent $\bm{x}_1$, which is compressed by the LCN into $\bm{x}_c$ and a compression-aware feature $\bm{f}_t$. The TP predicts the denoising entry $\tau$ from $\bm{x}_c$, while the TA converts $\bm{f}_t$ into the visual context $\bm{c}_v$. Conditioned on $\bm{c}_v$, the VideoDiT denoises $\bm{x}^{c}_{\tau}$, into $\hat{\bm{x}}_1$, which is decoded to reconstruct $\hat{\bm{V}}_1$.}
    \label{fig:framework}
\end{figure*}

Since training diffusion models directly in the pixel space is computationally expensive, a VAE is generally employed to encode the original visual data into a compact latent representation:
\begin{equation}
    \bm{x}_1=\mathcal{E}(\bm{V}_g),
\end{equation}
where $\bm{V}_g$ denotes the ground-truth video and $\mathcal{E}$ is the pretrained vision encoder. The subscript follows the flow-matching convention, where $t=0$ represents the noise distribution and $t=1$ represents the clean data distribution.

For video compression, the compression network (CN) encodes the $\bm{x}_1$ into a compressed representation $\bm{x}_c$:
\begin{equation}
\label{eq:xc}
    \bm{x}_c = \mathrm{CN}(\bm{x_1}) = \bm{x_1} + \bm{e},
\end{equation}
where $\bm{e}$ represents the distortion introduced by compression and quantization. Unlike video generation, where flow-matching models traverse the full probability path from Gaussian noise $\bm{x}_0\sim\mathcal{N}(\bm{0},\bm{I})$, the compressed latent retains rich texture information and provides a more informative initialization. We therefore start denoising at timestep $\tau$, with the corresponding latent $\bm{x}^c_{\tau}$ defined as follows.
\begin{equation}
\label{eq:xctau}
\bm{x}^c_{\tau} =\tau \bm{x}_c + (1-\tau)\bm{x}_0, \;\tau\in(0,1).   
\end{equation}

Based on \eqref{eq:xctau}, the denoising trajectory from the compressed initialization $\bm{x}^c_{\tau}$ to the clean latent $\bm{x}_1$ is defined as:
\begin{equation}
\label{eq:xct}
    \bm{x}^c_t=\frac{t-\tau}{1-\tau}\bm{x_1}+\frac{1-t}{1-\tau}\bm{x}^c_{\tau},\;t\in[\tau, 1].
\end{equation}
Differentiating \eqref{eq:xct} with respect to $t$ and substituting Eqs.~\eqref{eq:xc} and~\eqref{eq:xctau} gives the corresponding target velocity field:
\begin{equation}
\label{eq:vtao}
    \bm{v}_\tau=\bm{x}_1-\bm{x}_0-\frac{\tau}{1-\tau}\bm{e}=\bm{v}_{\mathrm{pre}}-\bm{v}_{\mathrm{ft}},
\end{equation}
where $\bm{v}_\mathrm{ft}=\frac{\tau}{1-\tau}\bm{e}$ is the compression-aware correction required during fine-tuning. A smaller $\lVert\bm{v}_{\mathrm{ft}}\rVert$ reduces the fine-tuning difficulty and generally improves reconstruction fidelity. However, decreasing $\tau$ increases the amount of Gaussian noise introduced into the initial state in \eqref{eq:xctau}, granting the generative prior greater freedom but potentially damaging the fidelity of the reconstructed video. Therefore, an appropriate $\tau$ is required to balance reconstruction fidelity
and realism.

We formulate timestep selection as maximizing the starting timestep while
penalizing the required fine-tuning correction. We can apply a Lagrangian relaxation method to derive the best timestep $\tau^*$:
\begin{equation}
\label{eq:optimization}
    \tau^*=
    \underset{\tau\in[0,1]}{\arg\max}
    \left[\tau-\lambda\frac{\tau}{1-\tau} \lVert\bm{e}\rVert\right]=1-\sqrt{\lambda\lVert\bm{e}\rVert},
\end{equation}
where $\lambda$ controls the trade-off between retaining realism and reducing the fine-tuning difficulty. This result shows that $\tau^*$ decreases monotonically with the distortion $\lVert\bm{e}\rVert$, which means more severely distorted inputs require an earlier denoising entry. If the maximum compression-aware correction achievable through fine-tuning is approximated by a constant $\epsilon$, the constraint becomes:
\begin{equation}
    \frac{\tau}{1-\tau} \lVert\bm{e}\rVert\leq\epsilon.
\end{equation}
The feasible solution for $\tau$ can then be derived in another closed form as:
\begin{equation}
\label{eq:tau*}
    \tau^*=\frac{\epsilon}{\epsilon+||\bm{e}||}\;.
\end{equation}
The fine-tuned flow model then denoises the compressed representation into a high-quality latent $\hat{\bm{x}}_1$ that achieves both high fidelity and realism.
\begin{equation}
\hat{\bm{x}}_1=\bm{x}_{\tau^*}^c+\int^1_{\tau^*} \bm{v}_\theta(\bm{x}_t)dt\;.
    \label{eq:flow-inference}
\end{equation}

\subsection{Proposed Framework Design}

The overall architecture of our network is illustrated in Fig. \ref{fig:framework}, which mainly consists of a latent compression network (LCN), temporal adaptor (TA), timestep predictor (TP) and video diffusion modules. Based on the pre-trained Wan2.1-VACE generative model \cite{vace}, our framework utilizes its VAE encoder to derive the clean latent representation $\bm{x}_1$. We adapt the DCVC-RT \cite{DCVC-RT} architecture into our LCN tailored for latent-space feature coding. For I-frames, an intra-frame compression module composed of depthwise convolutions and residual blocks ensures faithful reconstruction. For P-frames, an implicit temporal modeling scheme extracts temporal context via a streamlined feature extractor. By concatenating this context with the current latent along the channel dimension, the encoder-decoder jointly reduces spatio-temporal redundancy, ultimately yielding the reconstructed latent $\bm{x}_c$ and the feature $\bm{f}_t$:
\begin{equation}
   [\bm{x}_c, \bm{f}_t]= \mathrm{LCN}(\bm{x}_1).
\end{equation}

To predict the optimal denoising start-timestep, we design a lightweight $\mathrm{TP}$ based on the reconstructed latent $\bm{x}_c$. Specifically, $\bm{x}_c$ is first aggregated into a fixed-size feature vector via spatial adaptive pooling, ensuring flexibility across arbitrary input resolutions. The pooled representation is then processed through linear layers followed by a sigmoid activation function $\sigma(\cdot)$, which maps the output into the range $[0, 1]$. The predicted timestep $\tau$ is linearly scaled into a predefined search interval $[\tau_{\min}, \tau_{\max}]$:
\begin{equation}
\label{eq:tau}
    \tau=\tau_{\min}+(\tau_{\max}-\tau_{\min})\sigma(\mathrm{TP}(\bm{x}_c)).
\end{equation}
Subsequently, we obtain the initial denoising state $\bm{x}^c_{\tau}$ based on the predicted timestep $\tau$ according to \eqref{eq:xctau}.

More importantly, the compression-aware feature $\bm{f}_t$ is processed by a TA, consisting of 2D spatial and 3D temporal convolutional residual blocks, to boost feature capability and extract richer visual context:
\begin{equation}
    \bm{c}_v=\mathrm{TA}(\bm{f}_t).
\end{equation}
The extracted visual context $\bm{c}_v$ is then injected into the DiT through the VACE module as a condition to guide the network in predicting the velocity field $\bm{v}_{\theta}$ at each timestep $t$:
\begin{equation}
{\bm{v}}_{\theta}=\mathrm{DiT}(\bm{x}^c_{\tau}|\mathrm{VACE}(\bm{c}_v), t)\;.
    \label{eq:framework_inf}
\end{equation}

Finally, the denoised latent $\hat{\bm{x}}_1$ is reconstructed based on \eqref{eq:flow-inference}, which is then passed through the pre-trained VAE decoder $\mathcal{D}(\cdot)$ to yield the final compressed video $\hat{\bm{V}}_1$:
\begin{equation}
    \hat{\bm{V}}_1=\mathcal{D}(\hat{\bm{x}}_1).
\end{equation}

\subsection{Training Strategies}

The compressed distribution is initially far from the pretrained flow manifold, and the predictor can change the denoising trajectory before the refinement model learns a meaningful correction field. We therefore use three progressive stages. We first predefine a fixed denoising start timestep $\tau_0$ and jointly train the compression network and VACE module. To align the denoised output with the clean latent and accurately adapt the velocity field of the flow model, we employ the following latent-space loss function in the first stage:
\begin{equation}
\label{eq:stage1}
\mathcal{L}_{l}=R(\hat{\bm{y}})+\lambda_1\lVert\hat{\bm{x}}_1-\bm{x}_1\rVert^2_2+\mathbb{E}_{t, \bm{x}_c^{\tau_0}}\lVert\bm{v}_{\theta}-\bm{v}_{\tau_0}\rVert^2_2\;,
\end{equation}
where $\hat{\bm{y}}=g_a(\bm{x}_1)$ denotes the quantized representation produced by the compression encoder and is used to estimate the bitrate $R(\hat{\bm{y}})$. The target velocity $\bm{v}_{\tau_0}$ is computed according to \eqref{eq:vtao} and the coefficient $\lambda_1$ balances the rate and reconstruction terms. During training, we gradually decrease $\lambda_1$ over successive iterations to prevent an excessively low initial bitrate from destabilizing the denoising network.

Subsequently, we fine-tune the entire framework in the pixel space to further improve its end-to-end reconstruction capability. The pixel-level loss function is defined as
\begin{equation}
\label{eq:stage2}
    \begin{aligned}
        \mathcal{L}_{p}=R(\hat{\bm{y}})&+\lambda_2(\lVert\hat{\bm{x}}_1-\bm{x}_1\rVert^2_2+\lVert{\bm{x}}_c-\bm{x}_1\rVert^2_2\\
        &+\lVert\hat{\bm{V}}_1-\bm{V}_g\rVert^2_2
        +0.1\mathcal{L}_{\mathrm{LPIPS}}(\hat{\bm{V}}_1, \bm{V}_g))\;,
    \end{aligned}
\end{equation}
where the first two distortion terms enforce latent-space consistency for the denoised and compressed representations while the remaining two promote the fidelity and perceptual realism of the reconstructed videos. The coefficient $\lambda_2$ controls the trade-off between bitrate and reconstruction quality.

Finally, we introduce the timestep predictor for joint training to obtain the adaptive denoising entry timestep. Its parameters are initialized such that the initial prediction equals the predefined timestep $\tau_0$, ensuring a smooth transition and stable optimization. The predicted timestep is constrained to the interval $[\tau_{\min},\tau_{\max}]$ centered at $\tau_0$. We then continue optimizing the entire framework using \eqref{eq:stage2} to obtain the final timestep-adaptive model.

\subsection{Denoising Entry Analysis}

\begin{figure}[ht]
    \centering
    \includegraphics[width=1\linewidth]{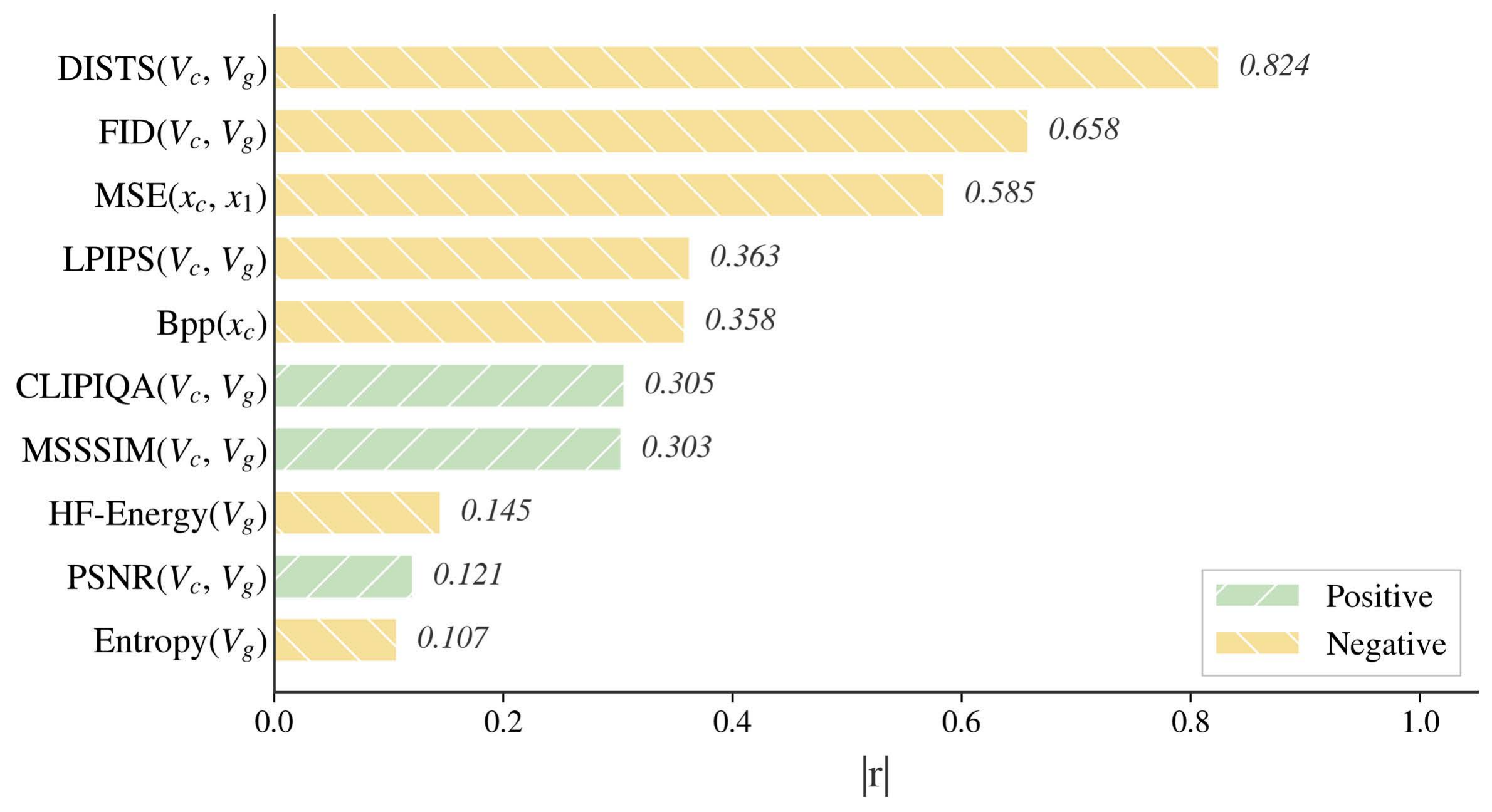}
    \caption{Pearson correlations between the predicted denoising start timestep $\tau$ and different distortion or content-related factors. DISTS exhibits the strongest negative correlation with $\tau$, whereas intrinsic source-video attributes show substantially weaker correlations.}
    \label{fig:correlation_bar}
\end{figure}

The trained ADE-GVC acquires the ability to predict the denoising start timestep.
According to \eqref{eq:tau*}, the optimal $\tau$ decreases monotonically with the latent distortion $\lVert\bm{e}\rVert$. To validate this relationship, we measure the distortion between $\bm{x}_c$ and $\bm{x}_1$ using latent-space MSE and various quality metrics between the directly decoded video $\bm{V}_c$ and the ground truth $\bm{V}_g$. These metrics include the pixel-level metrics PSNR, SSIM, and MS-SSIM, the perceptual metrics LPIPS and DISTS, the distribution-level metric FID, and the no-reference metrics NIQE and CLIPIQA. Since $\bm{x}_c$ is derived from $V_g$, $\tau$ may also be affected by the intrinsic characteristics of the source video. We therefore additionally examine brightness, contrast, entropy, Laplacian variance, color richness, edge density, high-frequency energy, sharpness, and compressed bitrate. These attributes characterize the visual complexity and information content of the source video. We calculate the Pearson correlation coefficient between each factor and the predicted denoising start timestep $\tau$. The top ten factors with the largest absolute correlation coefficients are reported in Fig.~\ref{fig:correlation_bar}.

The results show that DISTS exhibits the strongest correlation with $\tau$, achieving a coefficient of $-0.84$. We therefore use the decoded-domain DISTS distance $\rho_e$ as an observable proxy for the latent compression distortion $\lVert\bm{e}\rVert$:
\begin{equation}
    \rho_e=\mathrm{DISTS}\!\left(V_g,\mathcal{D}(\bm{x}_c)\right)
    \rightarrow
    \lVert\bm{e}\rVert.
    \label{eq:dists-distortion}
\end{equation}
In contrast, intrinsic video attributes and coding parameters exhibit much weaker correlations with $\tau$, indicating that the denoising entry is determined primarily by compression-induced distortion rather than source-video content.

\begin{figure}[t]
    \centering
    \includegraphics[width=0.96\linewidth]{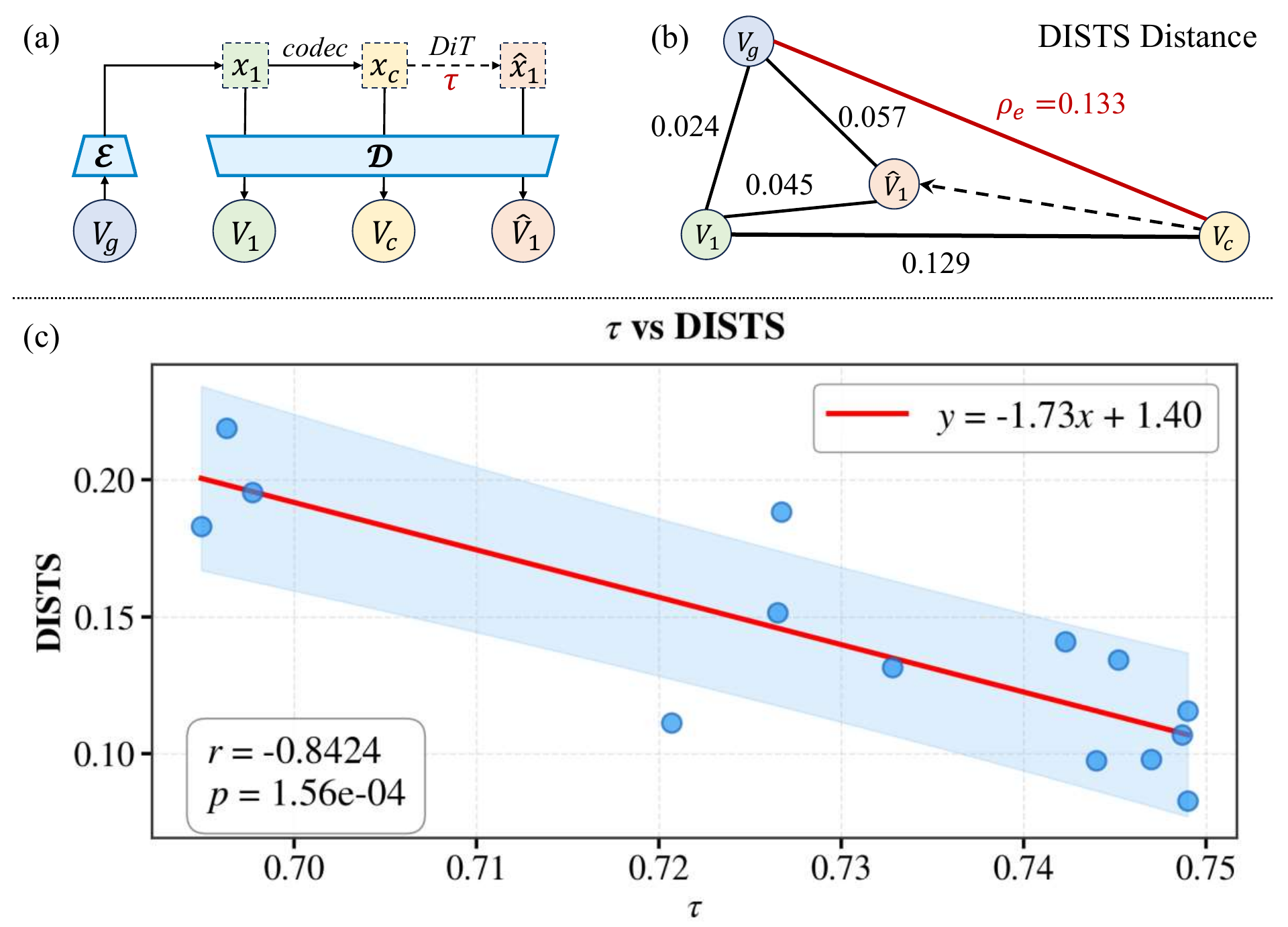}
    \caption{(a) Distortion evolution pipeline through the proposed framework. (b)  Distortion triangle in terms of the DISTS metric, which exhibits the strongest correlations with the predicted denoising start timestep.  (c) Relationship between the predicted $\tau$ and the distortion $\rho_e$. The strong negative correlation indicates that $\tau$ varies inversely with $\rho_e$, indicating that more severely distorted videos require an earlier denoising start timestep.}
    \label{fig:analysis}
\end{figure}

To further investigate the relationship between $\rho_e$ and the predicted denoising entry, we construct a distortion triangle based on $\rho_e$, as illustrated in Fig.~\ref{fig:analysis}. Specifically, DISTS is employed to quantify the perceptual distortion introduced at each stage of the framework, including compression and subsequent generative reconstruction. We then visualize the relationship between the compression-induced distortion $\rho_e$ and the predicted timestep $\tau$. From Fig.~\ref{fig:analysis}, a strong negative correlation is observed, indicating that videos with larger compression distortion tend to select an earlier denoising entry. Samples satisfying $\tau=\tau_{\max}$ are excluded from the correlation analysis because their predictions are saturated at the upper boundary and therefore do not accurately reflect the underlying distortion-dependent variation in $\tau$. This result empirically supports the relationship derived in \eqref{eq:tau*}.

\section{Experiments}

\subsection{Experimental Settings}

\paragraph{Datasets.}
Following \cite{li2026progvc}, we construct our training dataset from the Pexels dataset. We retain videos with high aesthetic and clarity scores, durations longer than 10 seconds, resulting in approximately 480K high-quality training videos. For evaluation, we conduct all experiments on widely used
benchmarks, including HEVC Class B~\cite{HEVC-B}, UVG~\cite{UVG}, and MCL-JCV~\cite{MCL-JCV} datasets.

\begin{figure*}[ht]
    \centering
    \includegraphics[width=0.96\linewidth]{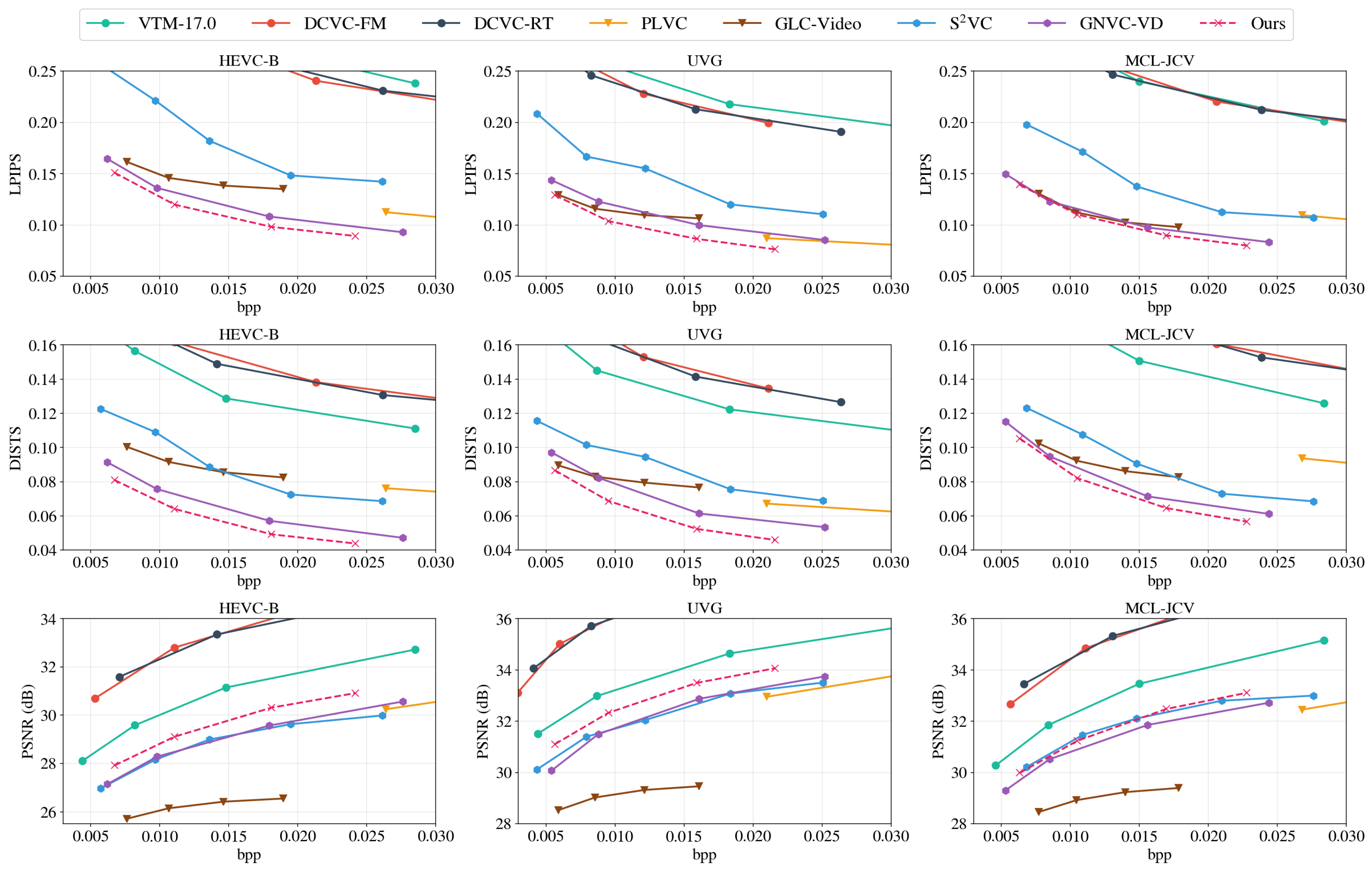}
    \caption{Quantitative comparisons with the state-of-the-art methods in terms of perceptual metrics (namely, LPIPS and DISTS) and supervised pixel-level metrics (namely, PSNR) on the HEVC-B, UVG, and MCL-JCV datasets.}
    \label{fig:RD_Curves}
\end{figure*}

\paragraph{Baselines and Metrics.}

We compare ADE-GVC with the conventional codec VTM-17.0~\cite{bross2021overview}, learned neural codecs DCVC-FM~\cite{DCVC-FM} and DCVC-RT~\cite{DCVC-RT}, and generative methods PLVC~\cite{PLVC}, GLC-Video~\cite{GLC}, GNVC-VD \cite{Mao_2026_CVPR}, and S$^2$VC~\cite{S2VC}. We evaluate the first 96 frames of each test sequence after conversion to the RGB color space. For closed-source methods such as GNVC-VD and S$^2$VC, we digitize the quantitative results from the rate-distortion curves reported in their papers. The bitrate is measured in bits per pixel (Bpp), while distortion is evaluated in terms of both reconstruction fidelity and perceptual quality. Specifically, we use PSNR to measure fidelity and LPIPS~\cite{LPIPS} and DISTS~\cite{DISTS} to assess perceptual quality.

\paragraph{Implementation Details}

We use Wan2.1 \cite{wan} as the pre-trained video diffusion backbone and incorporate conditioning adapter blocks following the design of VACE \cite{vace}. For training, in the first stage, we set $\tau_0$ to 0.7 and initialize $\lambda_1$ to 1024, halving it every 10K iterations. We train for 50K iterations with a batch size of 8, where each sample contains 13 frames. In the second stage, we train four bitrate models using $\lambda_2\in\{0.5,1,2,3\}$ for 100K iterations. The batch size is reduced to 4, each sample contains 25 frames, and the number of sampling steps is set to 5. In the third stage, we train the timestep predictor for 100K iterations and constrain its output to $\tau\in[0.65,0.75]$ to ensure stable optimization. During inference, each 96-frame sequence is divided into four groups of $\{25,25,25,21\}$ frames. All experiments are conducted on NVIDIA H20 GPUs.

\subsection{Experimental Results}

\paragraph{Quantitative Comparisons.}
We quantitatively evaluated the rate-distortion performance of ADE-GVC on the HEVC Class B~\cite{HEVC-B}, UVG~\cite{UVG}, and MCL-JCV~\cite{MCL-JCV} datasets. As shown in Fig.~\ref{fig:RD_Curves}, ADE-GVC consistently outperforms existing methods in both perceptual metrics and also substantially improves PSNR over other generative codecs. Using S$^2$VC~\cite{S2VC} as anchor, Table~\ref{tab:BD-rate} shows that ADE-GVC achieves average BD-rate savings of $62.97\%$ and $51.97\%$ in LPIPS and DISTS, significantly outperforming all competing methods. Compared with the recent open-source GLC-Video~\cite{GLC}, it further reduces the BD-rate by $21.54\%$, $40.99\%$, and $95.24\%$ in LPIPS, DISTS, and PSNR, demonstrating a favorable balance between perceptual quality and reconstruction fidelity.

\begin{table}[ht]
\centering
\resizebox{\linewidth}{!}{
\begin{tabular}{@{}lcccccc@{}}
\toprule
\multirow{2}{*}{Method/Metrics} & \multicolumn{2}{c}{HEVC-B} & \multicolumn{2}{c}{UVG} & \multicolumn{2}{c}{MCL-JCV}\\ \cmidrule(l){2-7} 
 & \multicolumn{1}{l}{LPIPS} & \multicolumn{1}{l}{DISTS} & \multicolumn{1}{l}{LPIPS} & \multicolumn{1}{l}{DISTS} & \multicolumn{1}{l}{LPIPS} & \multicolumn{1}{l}{DISTS} \\ \midrule
VVC &+248.8  &+195.6 &+442.6 &+390.8 &+348.7 &+364.7 \\
DCVC-FM &+248.2  &+542.1 &+371.9 &+804.0 &+375.9 &+688.4 \\
DCVC-RT &+277.1  &+568.7 &+392.7 &+801.4 &+409.8 &+740.0 \\
S$^2$VC &0.0 &0.0 &0.0 &0.0 &0.0 &0.0 \\
GLC-Video &-52.0  &-13.4 &-61.54 &-42.9 &-51.7 &-23.0 \\
GNVC-VD &-62.0  &-47.3 &-54.9 &-45.6 &-54.0 &-36.9 \\ \midrule
Ours &\textbf{-67.3}  &\textbf{-56.4} &\textbf{-67.4} &\textbf{-59.2} &\textbf{-54.2} &\textbf{-40.3} \\ \bottomrule
\end{tabular}
}
\caption{BD-Rate (\%) comparisons by anchoring S$^2$VC on three datasets. Bold values indicate the best results.}
\label{tab:BD-rate}
\end{table}

\begin{figure*}[ht]
    \centering
    \includegraphics[width=0.97\linewidth]{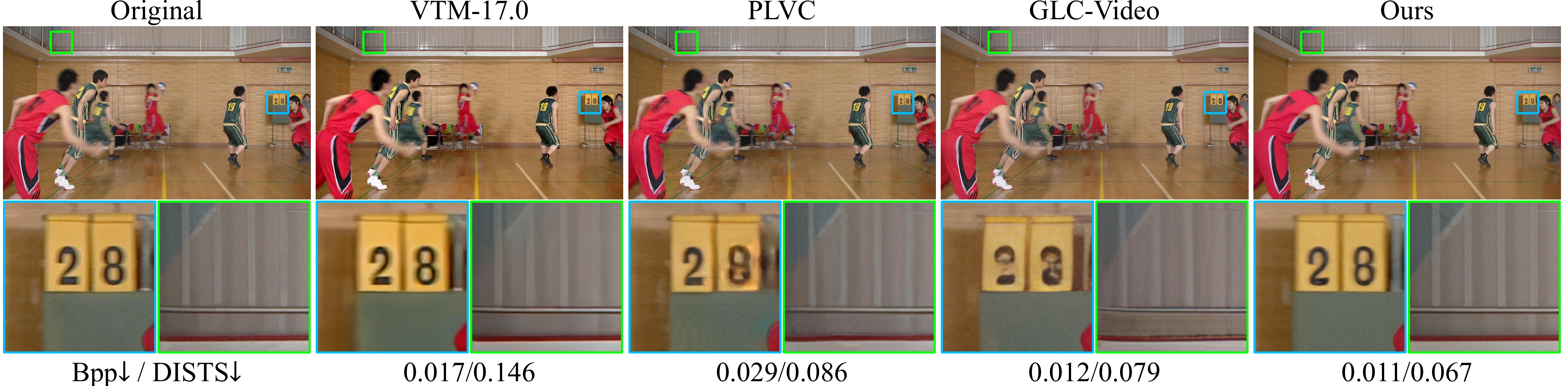}
    \caption{Qualitative comparisons on a single frame in terms of representative VVC, PLVC, GLC-Video and our ADE-GVC methods. We also report the Bpp/DISTS metrics to quantify compression efficiency.}
    \label{fig:sub}
\end{figure*}

\begin{figure}[ht]
    \centering
    \includegraphics[width=0.95\linewidth]{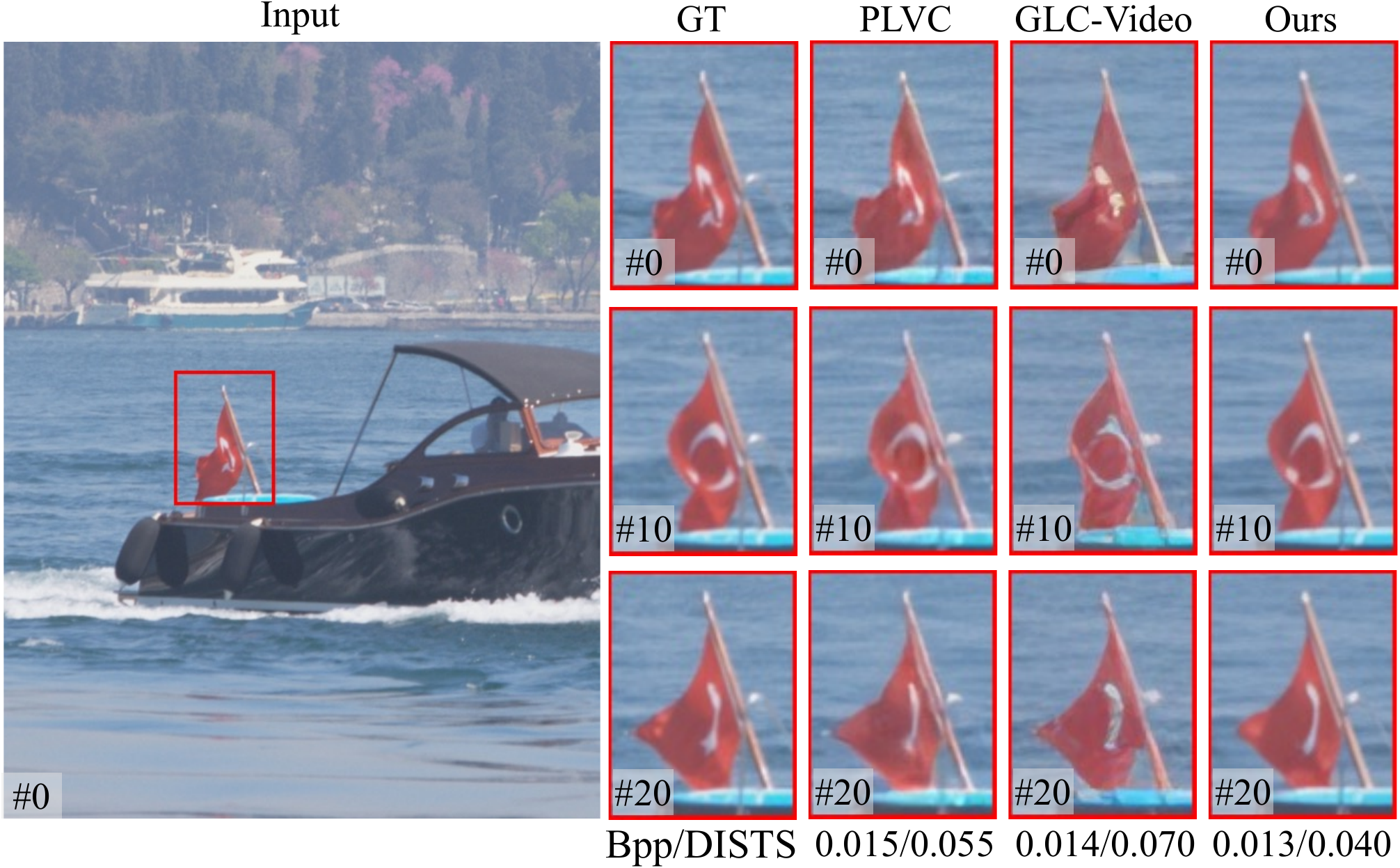}
    \caption{Qualitative comparisons for temporal consistency in terms of representative PLVC, GLC-Video and our ADE-GVC methods. We also report the Bpp/DISTS metrics to quantify compression efficiency.}
    \label{fig:sub-temp}
\end{figure}

\paragraph{Qualitative Comparisons.}
Fig.~\ref{fig:sub} presents a visual comparison of the reconstructed frames produced by the proposed method and the competing algorithms, with particular attention paid to textual and textured regions. The reconstruction produced by the conventional codec VTM-17.0 exhibits substantial over-smoothing, resulting in the loss of fine wall textures and blurred structural boundaries. Although existing generative video codecs improve perceptual realism, they remain prone to hallucinated or structurally inconsistent details. Specifically, PLVC introduces noticeable artifacts around the text, while GLC-Video severely distorts its global structure and reduces its readability. In contrast, ADE-GVC effectively avoids both the over-smoothing of conventional coding and the structural artifacts of existing generative methods, faithfully recovering delicate wall textures while preserving sharp, complete, and recognizable text boundaries. Fig.~\ref{fig:sub-temp} further demonstrates the temporal stability of ADE-GVC. For objects undergoing complex motion, even at extremely low bitrates, ADE-GVC largely avoids the temporally inconsistent textures and artifacts that are commonly introduced by other generative methods.

\begin{table}[ht]
    \centering
    \resizebox{\linewidth}{!}{\begin{tabular}{c|cccc}
    \toprule
    Optimization     & Baseline & +Parallelism  & +Compile & +DiT-bf16\\ \midrule
    Enc. Time(s) &0.306 &0.078 &0.078 &\textbf{0.078} \\
    Dec. Time (s) &4.341 &1.163 &1.133 &\textbf{0.831} \\
    Speedups  &1$\times$ &3.74$\times$ &3.84$\times$ &\textbf{5.11$\times$} \\ \bottomrule
    \end{tabular}}
    \caption{Complexity cumulative optimization for 1920x1080 video sequences, with runtime averaged over 96 frames. }
    \label{tab:complexity}
\end{table}

\subsection{Complexity Analysis}

Generative models are computationally intensive for high-resolution video reconstruction. To improve time efficiency, we progressively apply three optimization strategies, as shown in Table~\ref{tab:complexity}. The baseline is evaluated in FP32 on a single NVIDIA H20 GPU. Since the clips are processed independently, we distribute them for parallel inference, reducing the total coding time from 4.647s to 1.241s and achieving a $3.74\times$ speedup. Applying Torch compile function further reduces the decoding time to 1.133s per frame. Finally, BF16 inference for the DiT and VACE modules decreases the decoding time to 0.831s with negligible impact on reconstruction quality. Together, these optimizations achieve an overall speedup of $5.11\times$ at 1080p resolution.

\subsection{Ablation Studies}

We investigate the effects of progressive training and timestep selection in Table~\ref{tab:ablation}. Stage I, trained solely in the latent domain using \eqref{eq:stage1}, serves as the anchor. Stage II additionally applies the pixel-domain loss function in \eqref{eq:stage2} with a fixed starting timestep of $0.7$, substantially improving both fidelity and perceptual quality. Stage III introduces adaptive timestep prediction within the interval defined by \eqref{eq:tau}. With a wide range of $[0.60,0.90]$ (Stage III-Wide), the predicted $\tau$ values concentrate near $0.90$, shortening the generative refinement trajectory and yielding limited gains. Restricting the range to $[0.65,0.75]$ achieves a better balance between latent fidelity and generative refinement. Our final model consistently performs best, achieving BD-rate savings of up to $40.52\%$ on HEVC-B and $44.09\%$ on UVG with DISTS.

\begin{table}[ht]
\centering
\resizebox{\linewidth}{!}{
\begin{tabular}{@{}lcccccc@{}}
\toprule
\multirow{2}{*}{Method/BD-BR} & \multicolumn{3}{c}{HEVC-B}  & \multicolumn{3}{c}{UVG} \\ \cmidrule(l){2-4}\cmidrule(l){5-7} 
 & \multicolumn{1}{l}{LPIPS} & \multicolumn{1}{l}{DISTS} & \multicolumn{1}{l}{PSNR} & \multicolumn{1}{l}{LPIPS} & \multicolumn{1}{l}{DISTS} & \multicolumn{1}{l}{PSNR} \\ \midrule
Stage I &+0.0  &+0.0 &+0.0 &+0.0 &+0.0 &+0.0 \\
Stage II &-29.82& -33.65& -28.59& -30.27& -39.49& -29.87 \\
Stage III-Wide  &-31.71& -33.62& -30.44& -39.68& -40.92& -29.80 \\
Stage III-Ours &\textbf{-36.53}& \textbf{-40.52}& \textbf{-32.86}& \textbf{-44.48}& \textbf{-44.09}& \textbf{-33.89} \\ \bottomrule
\end{tabular}
}
\caption{Ablation study on the progressive training of our ADE-GVC on two datasets, evaluated by the LPIPS, DISTS, and PSNR metrics. Bold values indicate the best results.}
\label{tab:ablation}
\end{table}

\section{Conclusion and Future Work}

We presented ADE-GVC, a timestep-aware generative video compression framework that adaptively selects the denoising entry from the compressed latent. We formulated denoising entry selection as a geometric optimization problem under bounded fine-tuning capacity and showed that the optimal entry timestep decreases monotonically with the compression distortion. Therefore the fixed entry cannot accommodate the varying distortion levels of different video sequences. By combining a lightweight timestep predictor with compression-aware conditioning, ADE-GVC achieves a favorable balance between reconstruction fidelity and perceptual realism. Experiments demonstrate state-of-the-art performance and validate the relationship between the predicted timestep and DISTS. This work primarily investigates the selection of the denoising entry, while the number of sampling steps remains fixed across video sequences. Future work will jointly adapt the denoising entry and sampling steps to further improve reconstruction quality and inference efficiency. More broadly speaking, we believe that the denoising entry is an important yet overlooked design dimension for diffusion-based method, and that our analysis may also benefit generative image compression and other fidelity-critical low-level vision tasks.

{
    \small
    \bibliographystyle{ieeenat_fullname}
    \bibliography{main}
}


\end{document}